\documentclass[letterpaper, 10pt, conference]{ieeeconf}      

\IEEEoverridecommandlockouts                              %
\usepackage{subcaption}
\usepackage{algorithm}
\usepackage{algpseudocode}
\usepackage{amssymb}  
\usepackage{amsfonts}       
\usepackage{amsmath}
\usepackage{mathtools}
\usepackage{bm}
\usepackage{bbm}
\usepackage{changepage}
\usepackage[makeroom]{cancel}
\usepackage{sidecap}
\usepackage{longtable}
\usepackage{pifont}
\usepackage[dvipsnames]{xcolor}
\usepackage{textcomp,    
	xspace}
\usepackage{pgfplots}
\usepackage{tikz} 
\usepackage{bigints}
\usepackage{xcolor} 
\usepackage{colortbl}  

\usetikzlibrary{external}
\usetikzlibrary{pgfplots.groupplots}
\usepackage{hyperref}

\usepackage{dblfloatfix}
\title{\LARGE \bf
Dimensionality Reduction of Movement Primitives in Parameter Space
}

\author{Samuele Tosatto$^{1}$, Jonas Stadtm{\"u}ller$^{1}$ and Jan Peters$^{1,2}$
\thanks{*This work was supported by the Bosch Foschungsstiftung program.}
\thanks{$^{1}$Department of Computer Science,
        Technische Universit{\"a}t Darmstadt, 7500 Darmstadt, Germany.}%
\thanks{$^{2}$Department for Empirical Inference and Machine Learning,
	Max Planck Institute for Intelligent Systems, 7500 Stuttgart, Germany.
        Correspondence to 
        {\tt\small samuele.tosatto@tu-darmstadt.de}.}%
}

\newcommand{\de}{\mathrm{d}}

\algnewcommand\algorithmicforeach{\textbf{for each}}
\algdef{S}[FOR]{ForEach}[1]{\algorithmicforeach\ #1\ \algorithmicdo}

\begin{document}

\maketitle
\thispagestyle{empty}
\pagestyle{empty}

\begin{abstract}
Movement primitives are an important policy class for real-world robotics. 
However, the high dimensionality of their parametrization makes the policy optimization expensive both in terms of samples and computation.
Enabling an efficient representation of movement primitives facilitates the application of machine learning techniques such as reinforcement on robotics.
Motions, especially in highly redundant kinematic structures, exhibit high correlation in the configuration space.
For these reasons, prior work has mainly focused on the application of dimensionality reduction techniques in the configuration space. 
In this paper, we investigate the application of dimensionality reduction in the parameter space, identifying \textsl{principal movements}. The resulting approach is enriched with a probabilistic treatment of the parameters, inheriting all the properties of the Probabilistic Movement Primitives.
We test the proposed technique both on a real robotic task and on a database of complex human movements.
The empirical analysis shows that the dimensionality reduction in parameter space is more effective than in configuration space, as it enables the representation of the movements with a significant reduction of parameters.

\end{abstract}

\section{Introduction}

Robot learning is a promising approach to enable more intelligent robotics, which can easily adapt to the user's desires.
In recent years, the field of reinforcement learning (RL) has experienced an enormous advance in solving simulated tasks such as board- or video-games \cite{mnih_human-level_2015, lillicrap_continuous_2016, schulman_proximal_2017}, in contrast to little improvements in robotics. 
The major challenges in the direct application of RL to real robotics, are mainly the limited availability of samples and the fragility of the system, which disallow the application of unsafe policies.
These two disadvantages become even more evident when we consider that usual robotic tasks such as industrial manipulation, are defined in a high dimensional state and action space. A usual approach in the application of RL to robotics, is to initialize the policy via imitation learning \cite{schaal_is_1999,billard_robot_2004,argall_survey_2009,rana_towards_2018}.
In the past, there has been some effort in providing a safe representation of the policy for robotics, mainly by the means of Movement Primitives (MPs) \cite{davella_combinations_2003,schaal_learning_2005,khansari-zadeh_learning_2011,paraschos_probabilistic_2013}. The general framework of MPs has been extensvely studied and employed in a large variety of settings \cite{amor_interaction_2014,maeda_learning_2014,koert_demonstration_2016,maeda_active_2017,stark_experience_2019}.
\begin{figure}
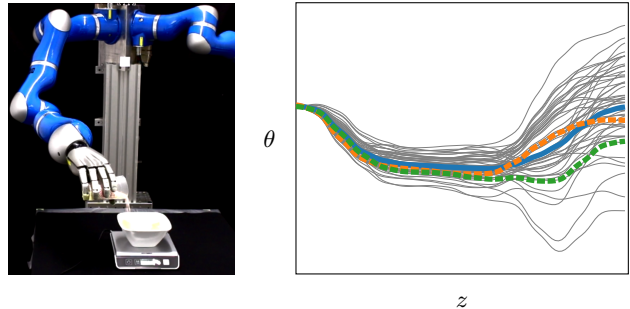

	\begin{subfigure}[t]{0.35\columnwidth}
		\input{plots/pictures/darias-pouring.tikz}
	\end{subfigure}
	\hspace{0.2cm}
	\begin{subfigure}[t]{0.63\columnwidth}
		\vspace{-3.74cm}
		\input{plots/analysis/example_primo.tikz}
	\end{subfigure}
	\caption{Our binamual platform performing a pouring task using Pro-Primos (left). 
		The generation of trajectories via PriMos (right): the blue line corresponds to the mean movement, while the dashed lines to the principal movements. By only linearly combining two principal movements, it is possible to generate a wide variety of movements (gray lines).
		\label{fig:first_page}}
\end{figure}
\begin{figure*}[t]
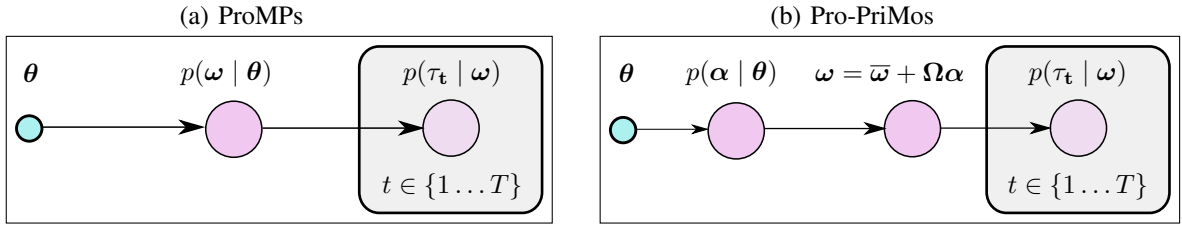

	\centering
	\begin{subfigure}[t]{0.9\columnwidth}
		\input{plots/graphs/promp.tikz}
	\end{subfigure}
	\begin{subfigure}[t]{0.9\columnwidth}
		\input{plots/graphs/pro-primos.tikz}
	\end{subfigure}
	\caption{Graphical model of the ProMPs framework (a) compared to Pro-PriMos (b). The main idea of Pro-PriMos is to generate a distribution of linear combinations of the principal-movements $\bm{\Omega}$.}
	\label{fig:graphical_models}
\end{figure*}
MPs have been shown to be effective when used for the direct application of RL to robotics \cite{peters_reinforcement_2008,kober_policy_2009,mulling_learning_2013}.
The drawback of the general framework of MPs, is the usual high number of parameters. In fact, in the configuration-space, the number of parameters is equal to the number of basis functions (typically greater than 10), times the number of degree-of-freedom (DoF). The number of parameters grows even quadratically w.r.t. the number of basis function and DoF when we want to represent the full covariance matrix in the case of Probabilistic Movement Primitives (ProMPs) \cite{paraschos_probabilistic_2013, paraschos_using_2018}.
In recent years, many authors have proposed techniques to solve the problem by using a latent representation of the robot's kinematic structure \cite{colome_dimensionality_2014-1,colome_dimensionality_2014,chen_efficient_2015,chen_dynamic_2016}. Especially for complex systems, such as humanoids, many DoF are redundant, and a compressed representation of the configuration-space results in a lighter parametrization of the MP.
However, this approach is less effective in the case of a robotic arm with fewer DoF Moreover, the parametrization of the MPs would be still linear w.r.t. the number of basis functions.

In this article, we build on the general framework of ProMPs and we propose to apply the dimensionality reduction in parameter space. In tasks where there is a high correlation between the movements, it makes more sense to seek a compressed representation of the movement instead of the configuration-space. To this end, we propose an approach where the movements can be seen as a linear combination of \textsl{principal movements} (Fig.\ref{fig:first_page}). We enrich our framework with a probabilistic treatment of the parameters, so that our approach inherits all the properties of ProMPs. 
We analyze the benefit of dimensionality reduction in the parameter space w.r.t. in configuration space, both on a challenging human motion reconstruction and in a robotic pouring task (Fig.~\ref{fig:first_page}). Our findings show that our proposed approach achieve satisfying accuracy even with a significant reduction of parameters. 
%

\section{Related Work}

The issue of dimensionality reduction in the context of motor primitives has been extensively studied.
To achieve dimensionality reduction for Dynamic Movement Primitives (DMPs) \cite{schaal_learning_2005}, the \textsl{autoencoded dynamic movement primitives} model  proposed in \cite{chen_efficient_2015}, uses to find a representation of the movements in a latent feature space, while in \cite{chen_dynamic_2016} the DMPs are embedded into the latent space of a time-dependent variational autoencoder.
In \cite{colome_dimensionality_2014} a linear projection in the latent space of the configuration space is considered, as well as the adaption of the projection matrix in the RL context. Interestingly, they consider the possibility to address the dimensionality reduction in the parameter space, but they discard this option as the projection matrix is more difficult to adapt since it is of higher dimensions. We agree with this argument, however, when we assume that the projection matrix can be considered fixed this argument is not valid anymore.
The dimensionality reduction of ProMPs has been addressed in \cite{colome_dimensionality_2018}, where the authors compare PCA versus an expectation-maximization approach in configuration space.
Until this point, all the literature is focused in finding a mapping between configuration space and a latent space, and proposing the learning of the MPs in this lower representation.
However, this approach results to be more efficient in complex kinematic structures, such as the human body, where the high number of joints facilitate the possibility of redundancies in the configuration space. Moreover, this reduction, does not affect the intrinsic high-dimensional nature of the MPs, which requires usually a high number of basis functions.

A way to overcome this problem is to focus in the parameter space of the movement primitives. The dimensionality reduction in this case exploits similarities between movements, and high correlations between parameters.
To the best of our knowledge, the only work performing a reduction in parameter is \cite{rueckert_extracting_2015}. The proposed setup is however complex as it considers a fully hierarchical Bayesian setting, where the movements are encoded by a mixture of Gaussian models. They learn those parameters using variational inference. 
Furthermore, their approach does not address the question of whether the parameter space reduction is more convenient or not.

In our paper, we want to focus on the comparison between the dimensionality reduction in configuration and parameter space, arguing that the latter is more convenient. 
To this end, we propose the Principal Movements (PriMos) framework, which enables the selection of \textsl{principal movements} and the subsequent representation of the movements in this convenient space. We extend PriMos to incorporate a probabilistic treatment of the parameters (Pro-PriMos), in a similar way to the ProMPs framework. Our approach only adds a linear transformation to the framework already developed by Parachos et al., as depicted in Fig.~\ref{fig:graphical_models}, therefore it maintains all the properties exposed by the ProMPs, such as time modulation, movements co-activation or movement conditioning.
To maintain the method simple, we select PCA for the dimensionality reduction, even if more sophisticated techniques can be used.\vspace{-1em}

\section{The Principal Movement Framework}

Machine learning, should allow the robotic agent to interact with the real world in a non-predetermined way. However, the possible generated movement should be smooth, and possibly constrained to be safe (not colliding with other objects, without excessive speed or accelerations, etc). These issues require a whole field of research to be solved. MPs are useful in this context.
The idea behind MPs, is to restrict the class of all possible robotic movements to a specific parameterized class usually by linearly combining a set of parameters with a set of features. For a particular class of features (e.g., radial basis function), the movements are guaranteed to be smooth.
However, sometimes restricting the space of robotic movement to be smooth is not enough, and we want also the movements to be distributed similarly to some demonstration provided by a human expert. The ProMPs provides a probabilistic treatment of the movement's parameters \cite{paraschos_probabilistic_2013, paraschos_using_2018}.
Both frameworks are useful and can be coupled with RL. 
In the following, we introduce the Principal Movements (PriMos) framework, as well as its probabilistic extension (Pro-PriMos). In Section~\ref{sec:preliminaries}, we introduce the formal notation of the MPs framework. In Sections~\ref{sec:primos} and \ref{sec:proprimos}, we introduce PriMos and Pro-PriMos relying on the assumption that a set of \textsl{principal movement} is given. In Section~\ref{sec:primos_pca} we propose a technique to find the \textsl{principal movements}, offering a complete algorithm for dimensionality reduction for MPs.

\subsection{Preliminaries}
\label{sec:preliminaries}
Let us concisely introduce the notation and formalize the classic MPs framework.
For simplicity, we will first consider only a one-dimensional trajectory $\bm{\tau}^\intercal = [\tau_{1}, \dots, \tau_{T}]$\footnote{Contrarily to \cite{paraschos_using_2018}, we do not consider the velocities, however they can be incorporated, without any loss of generality.}, where $\tau_i$ represent the joint's position at time $t_i$.
In order to introduce the \textsl{time modulation}, we use a \textsl{phase}-vector $\textbf{z}$ where $z_i = (t_i-t_1)/(t_T-t_1)$. 
Moreover, we consider $n$ normalized radial-basis functions $\Phi_i:\mathbb{R}\to \mathbb{R}$, usually, ordered to evenly cover the phase-space, they are centered between $[-2h, 1+2h]$ where $h$ is the bandwidth. At time $t$ the features are described by a $n$-dimensional column vector $\bm{\Phi}_t$ where $\Phi_{t, i} = \Phi_i(z_t)$.
Assuming that the observations are perturbed by a zero-mean Gaussian noise $\epsilon_t$ with variance $\sigma_\tau$, 
we want to represent the movement  $\tau_t = \bm{\Phi}_t^\intercal\bm{\omega} + \epsilon_t$ as a linear combination of parameters $\bm{\omega} \in \mathbb{R}^{n}$. , we obtain the following \textsl{maximum-likelihood estimation} (MLE) problem
%
\begin{equation}
\max_{\omega} \mathcal{L}(\bm{\omega}) = \max_{\omega}\prod_{t} \mathcal{N}(\tau_t \mid \bm{\Phi}_t^\intercal \omega, \sigma_t) \label{eq:frequentist_view}.
\end{equation}    
Equation \eqref{eq:frequentist_view} can be solved with usual linear regression, or more commonly with Ridge-regression (which corresponds to have a prior Gaussian distribution on $\omega$, $\mathcal{N}(\omega_i|0; \sigma_\omega)$) that leads to better generalization and is more numerically stable,
\begin{equation}
	\bm{\omega} = \left(\bm{\Phi}^\intercal\bm{\Phi}+ \lambda \mathbb{I}\right)^{-1}\bm{\Phi}^\intercal\bm{\tau} \label{eq:mpparams}
\end{equation}
where $\lambda$ is the ridge penalization term and $\bm{\Phi}^\intercal = [\bm{\Phi}_1,\bm{\Phi}_2 \dots, \bm{\Phi}_T]$. 
However, there is the possibility to extend this setting with a probabilistic treatment of the movement parameters.
In this case, we want to find the best distribution over parameters which encode a distribution of trajectories, i.e.,
\begin{equation}
p(\bm{\tau}\mid \mu_{\omega}, \Sigma_\omega) = \int p(\bm{\tau} \mid \omega)\mathcal{N}(\bm{\omega} \mid \mu_\omega; \Sigma_\omega) \de \bm{\omega}.
\end{equation}
This is the setting described with ProMPs in \cite{paraschos_probabilistic_2013}. For a finite set of stroke-movements $\{\bm{\tau}_i\}_{i=1}^m$, it is possible to use the parameters $\bm{\omega}_i$ estimated for each trajectory $\bm{\tau}_i$ in order to estimate $\bm{\mu}_{\omega}$ and $\bm{\Sigma}_\omega$, i.e., 
\begin{equation}
	\bm{\mu}_\omega = \frac{1}{m}\sum_{i=1}^m \bm{\omega}_i , \quad \bm{\Sigma}_\omega = \frac{1}{m}\sum_{i=1}^m (\bm{\omega}_i-\bm{\mu}_\omega)(\bm{\omega}_i-\bm{\mu}_\omega)^\intercal .
	\label{eq:prompparams}
\end{equation}
In a more realistic scenario, we need to represent multiple joints. Assuming a system with $d$ joints, we can encode $\bm{\tau}^\intercal = [\tau_{1,1}, \dots, \tau_{1,T}, \dots, {\tau_{d, T}}]$ where $\tau_{i, j}$ is the position of the $i^\text{th}$ joint at time $t_j$, $\bm{\Psi} = \mathbb{I} \otimes \bm{\Phi}$ (where $\mathbb{I}$ is the $d\times d$ identity matrix) and $\bm{\omega}$ column-vector of length $nd$ encoding the movement's parameters. Similarly to the notation already used, we denote with $\bm{\Psi}_t$ the feature matrix corresponding to the time step $t$.

\subsection{Principal Movements}
\label{sec:primos}
The MPs and ProMPs frameworks, usually require a large number of parameters for encoding the movements. In MPs, we need in fact $d \times n$ parameters (where $d$ is the dimension of the considered movement and $n$ is the number of radial basis functions). rThe choice of the number of radial-basis functions usually depends by both the speed and the complexity of the movements, but most of the time it is greater than ten. For a $7$ d.o.f. we, therefore, need at least $70$ parameters to encode the MPs. This makes the application of RL challenging for a 7 d.o.f. robot arm.
We will assume, from now on, that an oracle (i.e., a human expert or a dimensionality reduction method) will give us the parameters of a \textsl{mean-movement} $\overline{\bm{\omega}}$, and a matrix of \textsl{principal-movements} $\bm{\Omega}^\intercal = [\bm{\omega}^p_1, \dots, \bm{\omega}^p_{n_c}]$.
We therefore want to encode a given trajectory $\bm{\tau}$ as a linear combination of the principal movements, i.e.,
\begin{equation}
\bm{\tau}_t = \bm{\Psi}_t\overline{\bm{\omega}} + \bm{\Psi}_t\bm{\Omega}\bm{\alpha} + \epsilon_t . \label{eq:deftauprimo}
\end{equation}   
We consider $\bm{\alpha} \in \mathbb{R}^{n_c}$ the new parameter vector. Note that $n_c$ is generally independent of the number of joints $d$ and of the number of radial basis function $n$, but instead the choice of $n_c$ is connected to the \textsl{complexity} of the movement-space that we aim to represent.
The MLE problem
\begin{equation}
\max_{\bm{\alpha}} \prod_{t} \mathcal{N}(\bm{\tau}_t \mid \bm{\Psi}_t\overline{\bm{\omega}} + \bm{\Psi}_t\bm{\Omega}\bm{\alpha}, \sigma_t), \label{eq:primo_problem}
\end{equation}  
induced by \eqref{eq:deftauprimo} has a Ridge regression solution 
\begin{equation}
\bm{\alpha} = \left(\bm{\Omega}^\intercal\bm{\Psi}^\intercal\bm{\Psi}\bm{\Omega} + \lambda \mathbb{I}\right)^{-1}\bm{\Omega}^\intercal\bm{\Psi}^\intercal\left(\bm{\tau} - \bm{\Psi}\overline{\bm{\omega}}\right).
\label{eq:primosolution}
\end{equation}

\subsection{Probabilistic Principal Movements}
\label{sec:proprimos}
\begin{figure*}[t]
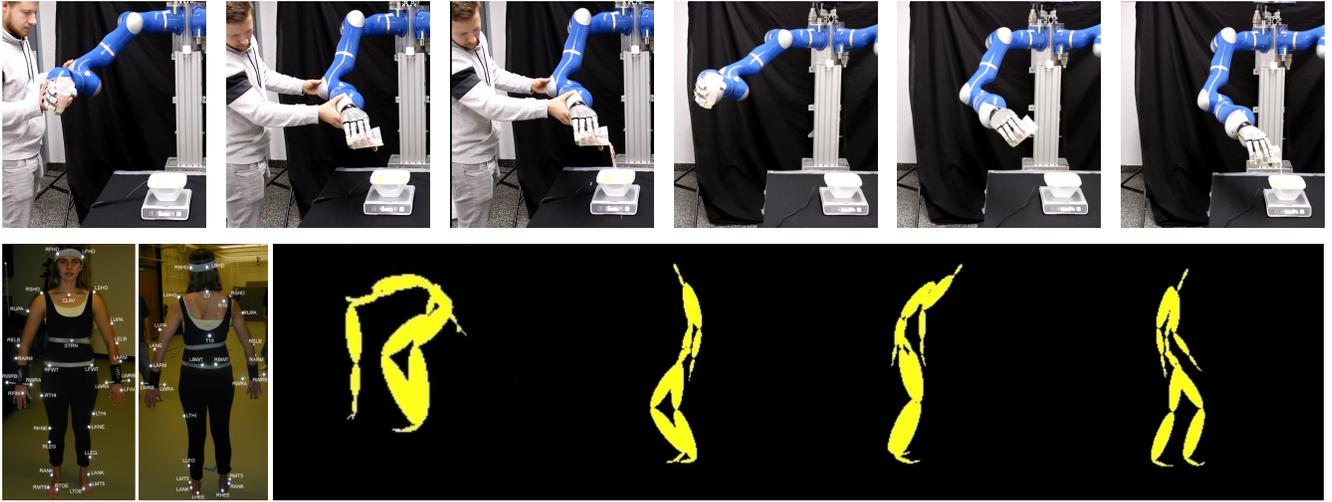

	\begin{subfigure}[t]{0.33\columnwidth}
		\input{plots/pictures/demo2.tikz}
	\end{subfigure}
	\begin{subfigure}[t]{0.33\columnwidth}
		\input{plots/pictures/demo3.tikz}
	\end{subfigure}
	\begin{subfigure}[t]{0.33\columnwidth}
		\input{plots/pictures/demo4.tikz}
	\end{subfigure}
	\begin{subfigure}[t]{0.33\columnwidth}
		\input{plots/pictures/primo1.tikz}
	\end{subfigure}
	\begin{subfigure}[t]{0.33\columnwidth}
		\input{plots/pictures/primo2.tikz}
	\end{subfigure}
	\begin{subfigure}[t]{0.33\columnwidth}
		\input{plots/pictures/primo3.tikz}
	\end{subfigure}
	
	\begin{subfigure}[t]{0.195\columnwidth}
		\input{plots/pictures/human_front.tikz}
	\end{subfigure}
	\begin{subfigure}[t]{0.192\columnwidth}
		\input{plots/pictures/human_back.tikz}
	\end{subfigure}
	\begin{subfigure}[t]{0.4\columnwidth}
		\input{plots/pictures/human_1.tikz}
	\end{subfigure}
	\begin{subfigure}[t]{0.39\columnwidth}
		\input{plots/pictures/human_2.tikz}
	\end{subfigure}
	\begin{subfigure}[t]{0.388\columnwidth}
		\input{plots/pictures/human_3.tikz}
	\end{subfigure}
	\begin{subfigure}[t]{0.39\columnwidth}
		\input{plots/pictures/human_4.tikz}
	\end{subfigure}
	\caption{Top: From left to right, the demonstration in kinestatic teaching mode of the pouring task, and the subsequent reconstruction of the movement with PriMos. Bottom: Markers placements on the human body, and illustrative example of the skeleton reconstruction for a ``walk and pick-up'' task of subject \#143 of the \texttt{MoCap} database \url{http://mocap.cs.cmu.edu}.}
	\label{fig:frames}
\end{figure*} 

A probabilistic treatments of the MPs, is often convenient, as it enables the application of statistical tools \cite{paraschos_probabilistic_2013, paraschos_using_2018}.
To enrich our approach with a probabilistic treatment, we assume our parameter vector $\bm{\alpha}$ to be multivariate-Gaussian distributed, i.e., $\bm{\alpha} \sim \mathcal{N}(\cdot \mid \bm{\mu}_{\alpha}; \bm{\Sigma}_\alpha)$.
Very similarly to the classic ProMPs approach, we have that
\begin{align}
& p(\bm{\tau}_t \mid \bm{\mu}_\alpha, \bm{\Sigma}_\alpha) \nonumber \\
=& \int \mathcal{N}(\bm{\tau}_t\mid \bm{\Psi}_t( \bm{\Omega}\bm{\alpha} + \overline{\bm{\omega}}), \bm{\Sigma}_\tau) \nonumber 
\mathcal{N}(\bm{\alpha} \mid \bm{\mu}_\alpha, \bm{\Sigma}_\alpha) \de \bm{\alpha} \\
= & \mathcal{N}(\bm{\tau}_t\mid \bm{\Psi}_t( \bm{\Omega}\bm{\mu}_{\alpha} + \overline{\bm{\omega}}), \bm{\Psi}_t\bm{\Omega}\bm{\Sigma}_\alpha\bm{\Omega}^\intercal  \bm{\Psi}_t^\intercal + \bm{\Sigma}_\tau).
\end{align}
The mean $\bm{\mu}_\alpha$ and the covariance $\bm{\Sigma}_\alpha$ can be estimated as similarly done for ProMPs,  (Eq.~\ref{eq:prompparams})
\begin{equation}
\bm{\mu}_\alpha = \frac{1}{m}\sum_{i=1}^m \bm{\alpha}_i , \quad \bm{\Sigma}_\alpha = \frac{1}{m}\sum_{i=1}^m (\bm{\alpha}_i-\bm{\mu}_\alpha)(\bm{\alpha}_i-\bm{\mu}_\alpha)^\intercal ,
\label{eq:proprimoparams}
\end{equation}
where the parameters $\bm{\alpha}_i$ correspond to the trajectories $\bm{\tau}_i$.

We assume an affine transformation between the full parameters $\bm{\omega}$ and the reduced $\bm{\alpha}$, as shown in Fig.~\ref{fig:graphical_models}, and $\bm{\alpha}$ is assumed to be Gaussian distributed. Under these assuptions, $\bm{\omega}$ is also Gaussian distributed, allowing for a mapping from Pro-PriMos to ProMPs,  
\begin{equation}
	\hat{\bm{\mu}}_\omega = \bm{\Omega}\bm{\mu}_{\alpha} + \overline{\bm{\omega}}, \quad \quad \hat{\bm{\Sigma}}_\omega =\bm{\Omega}\bm{\Sigma}_\alpha\bm{\Omega}^\intercal. \nonumber
\end{equation}
Hence, the proposed framework enjoys all the properties of the ProMPs, such as movements co-activation or movement conditioning.

\subsection{Inference of the Principal Components}
\label{sec:primos_pca}

Until this point, we considered the parameters $\overline{\bm{\omega}}$ and $\bm{\Omega}$ given by an oracle. We mentioned that the underlying assumption of our work is that each movement can be seen as a linear combination of some movements which we call \textsl{principal movements}. To be precise, we assume that every trajectory $\bm{\tau}$ can be seen as $\bm{\tau} = \overline{\bm{\tau}} + \sum_i \bm{\tau}^p_i\alpha_i$. Since there exists a linear combination between the trajectory $\bm{\tau}$ and its parameter $\bm{\omega}$, we can argue, that the same relation exists in the parameter space. Reasoning in the parameter space is more convenient since the trajectories can have different lengths, depending on the sampling frequency and duration. The parameter space is instead fixed. 
We can say that the parameter space is \textsl{frequency and duration agnostic}.
There are many techniques that can be used to estimate $\overline{\bm{\omega}}$ and $\bm{\Omega}$ when they are unknown, but among these, given the Gaussian assumption very made and also thanks to its simplicity, the Principal Component Analysis (PCA) \cite{pearson_lines_1901} seems to be the most suited. We can think of the movement being Gaussian distributed with a certain mean trajectory $\overline{\bm{\tau}}$ (with parameters $\overline{\bm{\omega}}$) and with some orientation. In its geometric interpretation, the PCA extracts the main axis of the covariance matrix (i.e. the Eigenvectors of the covariance), and we can express each point in the new space as a linear combination of the Eigenvectors. Even though the Singular Value Decomposition is a more efficient technique to perform the PCA \cite{golub_matrix_2012}, we continue with the Eigenvector decomposition to maintain this parallelism. 
Let $\{\mathbf{v}_i\}_{i=1}^{nd}$ be the Eigenvectors of $\bm{\Sigma}_\omega$ and $\lambda_i$ the corresponding Eigenvalues (with $|\alpha_1| \geq |\alpha_2| \geq \dots$).
To compute the projection matrix $\bm{\Omega}$ we select the first most informative $n_c$ Eigenvectors and we multiply them by the square roots of their corresponding Eigenvalues
\begin{equation}
	\bm{\Omega} = \left[\bm{v}_1\sqrt{|\lambda_1|}, \bm{v}_2\sqrt{|\lambda_2|}, \dots, \bm{v}_{n_c}\sqrt{|\lambda_{n_c}|} \right].\label{eq:pcaomega}
\end{equation}
This re-scaling gives us the possibility to have the \textsl{principal movements} scaled according to the variance of the data, and results in standardized values of $\bm{\alpha}$ (i.e. $\bm{\alpha} \sim \mathcal{N}(\cdot \sim \bm{0}, \mathbb{I})$).
Pro-PriMos is concisely summarized in Algorithm~\ref{alg:pro-primos}.

\begin{algorithm}[t]
	\caption{Pro-PriMos \label{alg:pro-primos}}
	\begin{algorithmic}
		\Require A dataset of trajectories $\{\mathbf{z}_i, \bm{\tau}_i\}_{i=1}^m$
		\ForEach {$\mathbf{z}_i, \bm{\tau}_i$}
			\State Compute $\bm{\omega}_i$ with \eqref{eq:mpparams}
		\EndFor
		\State Estimate $\bm{\mu}_\omega$ and $\bm{\Sigma}_\omega$ with \eqref{eq:prompparams}
		\State Compute $\bm{\Omega}$ with \eqref{eq:pcaomega}
		\State Set $\overline{\bm{\omega}} := \bm{\mu}_\omega$
		\ForEach {$\mathbf{z}_i, \bm{\tau}_i$}
			\State Compute $\bm{\alpha}_i$ with \eqref{eq:primosolution}
		\EndFor
		\State \Return $\bm{\mu}_\alpha\!=\!\frac{\sum_i\bm{\alpha}_i}{n_c}$; $\bm{\Sigma}_\alpha\!=\frac{\!\sum_i(\bm{\alpha}_i-\bm{\mu}_\alpha)(\bm{\alpha}_i-\bm{\mu}_\alpha)^\intercal}{n_c}$ 
	\end{algorithmic}
\end{algorithm}

\section{Empirical Analysis}
\begin{figure*}[t]
	\hspace{-1.5cm}
	\begin{subfigure}[t]{0.66\columnwidth}
		\input{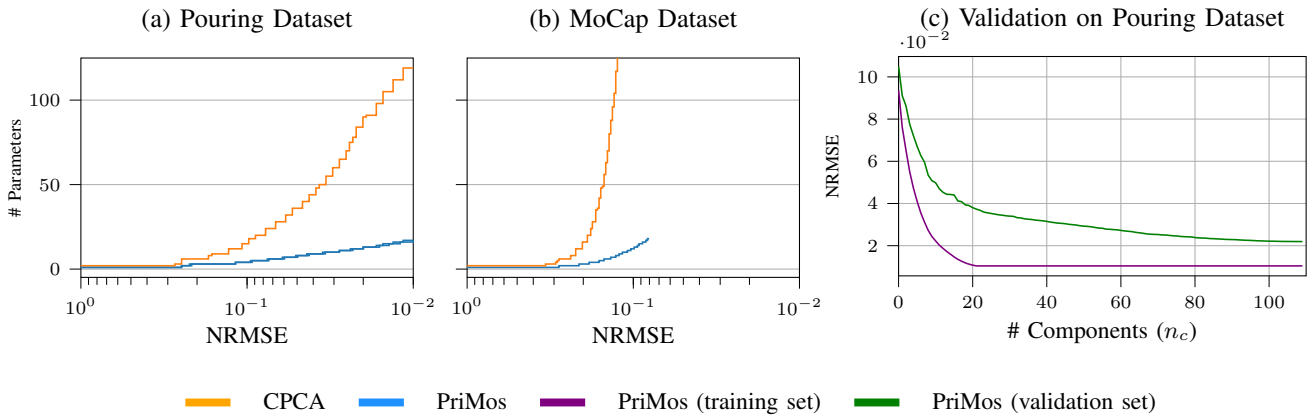}
	\end{subfigure}
	\begin{subfigure}[t]{0.57\columnwidth}
\begin{tikzpicture}

\definecolor{color0}{rgb}{0.12156862745098,0.466666666666667,0.705882352941177}
\definecolor{color1}{rgb}{1,0.498039215686275,0.0549019607843137}

\begin{axis}[
width=6cm,
height=4.5cm,
legend cell align={left},
legend style={fill opacity=0.8, draw opacity=1, text opacity=1, at={(0.03,0.97)}, anchor=north west, draw=white!80!black},
log basis x={10},
tick align=outside,
tick pos=left,
title={(b) MoCap Dataset},
x dir=reverse,
x grid style={white!69.0196078431373!black},
xlabel={NRMSE},
xlabel style={font=\small},
xticklabel style={font=\scriptsize},
xmin=0.01, xmax=1,
xmode=log,
xtick style={color=black, font=\scriptsize},
y grid style={white!69.0196078431373!black},
yticklabel style={font=\scriptsize},
ymajorgrids,
ymin=-4.9, ymax=124.9,
yminorgrids,
ytick style={color=black},
yticklabel=\empty
]
\addplot [semithick, color0, const plot mark right]
table {%
0.01 nan
0.0102341140210545 nan
0.0104737089795945 nan
0.0107189131920513 nan
0.0109698579789238 nan
0.0112266777351081 nan
0.0114895100018731 nan
0.0117584955405216 nan
0.0120337784077759 nan
0.0123155060329283 nan
0.0126038292967973 nan
0.0128989026125331 nan
0.0132008840083142 nan
0.0135099352119803 nan
0.0138262217376466 nan
0.0141499129743458 nan
0.0144811822767453 nan
0.0148202070579886 nan
0.0151671688847092 nan
0.0155222535742705 nan
0.0158856512942805 nan
0.0162575566644379 nan
0.0166381688607613 nan
0.017027691722259 nan
0.0174263338600965 nan
0.0178343087693191 nan
0.0182518349431904 nan
0.0186791359902078 nan
0.019116440753857 nan
0.0195639834351706 nan
0.0200220037181558 nan
0.0204907468981585 nan
0.0209704640132323 nan
0.021461411978584 nan
0.0219638537241655 nan
0.0224780583354873 nan
0.0230043011977292 nan
0.0235428641432242 nan
0.0240940356023952 nan
0.024658110758226 nan
0.0252353917043477 nan
0.0258261876068268 nan
0.0264308148697411 nan
0.0270495973046313 nan
0.0276828663039207 nan
0.0283309610183932 nan
0.0289942285388288 nan
0.0296730240818887 nan
0.0303677111803546 nan
0.0310786618778201 nan
0.0318062569279412 nan
0.0325508859983506 nan
0.0333129478793467 nan
0.0340928506974681 nan
0.0348910121340677 nan
0.0357078596490046 nan
0.0365438307095726 nan
0.037399373024788 nan
0.0382749447851631 nan
0.0391710149080926 nan
0.0400880632889846 nan
0.0410265810582719 nan
0.0419870708444391 nan
0.0429700470432084 nan
0.0439760360930272 nan
0.045005576757005 nan
0.0460592204114511 nan
0.0471375313411672 nan
0.0482410870416537 nan
0.04937047852839 nan
0.0505263106533568 nan
0.0517092024289676 nan
0.0529197873595844 nan
0.0541587137807947 nan
0.0554266452066311 nan
0.0567242606849198 nan
0.058052255160949 nan
0.0594113398496503 nan
0.0608022426164942 nan
0.0622257083673023 nan
0.0636824994471859 nan
0.0651733960488242 nan
0.0666991966303012 nan
0.0682607183427239 nan
0.0698587974678525 nan
0.0714942898659758 nan
0.073168071434272 nan
0.0748810385759002 nan
0.0766341086800746 nan
0.0784282206133768 nan
0.0802643352225717 18
0.0821434358491943 18
0.0840665288561833 17
0.086034644166845 16
0.0880488358164346 16
0.0901101825166502 15
0.0922197882333433 15
0.0943787827777538 14
0.096588322411587 13
0.0988495904662559 13
0.101163797976621 12
0.103532184329566 12
0.105956017927762 11
0.108436596868961 11
0.110975249641207 10
0.113573335834311 10
0.116232246867985 10
0.118953406737032 9
0.121738272773966 9
0.124588336429501 8
0.127505124071301 8
0.13049019780144 7
0.13354515629299 7
0.136671635646201 7
0.139871310264724 6
0.143145893752348 6
0.146497139830729 6
0.149926843278605 6
0.153436840893001 5
0.157029012472938 5
0.160705281826164 5
0.164467617799466 4
0.168318035333096 4
0.172258596539879 4
0.176291411809595 4
0.180418640939207 4
0.184642494289554 4
0.188965233969121 3
0.193389175045523 3
0.197916686785356 3
0.202550193923067 3
0.207292177959537 3
0.212145178491063 3
0.21711179456945 2
0.222194686093952 2
0.227396575235793 2
0.232720247896041 2
0.238168555197616 2
0.243744415012222 2
0.249450813523032 2
0.255290806823952 2
0.261267522556333 2
0.267384161583995 2
0.273643999707467 2
0.280050389418363 2
0.286606761694825 1
0.293316627839004 1
0.300183581357559 1
0.307211299886176 1
0.31440354715915 1
0.321764175025074 1
0.329297125509715 1
0.337006432927193 1
0.344896226040576 1
0.352970730273065 1
0.361234269970943 1
0.369691270719503 1
0.378346261713193 1
0.387203878181256 1
0.396268863870148 1
0.405546073584083 1
0.415040475785048 1
0.42475715525369 1
0.434701315812502 1
0.444878283112759 1
0.455293507486695 1
0.465952566866468 1
0.476861169771447 1
0.488025158365443 1
0.499450511585514 1
0.511143348344017 1
0.523109930805626 1
0.535356667741072 1
0.547890117959394 1
0.560716993820546 1
0.57384416483024 1
0.587278661318948 1
0.601027678207038 1
0.61509857885805 1
0.629498899022189 1
0.644236350872137 1
0.659318827133355 1
0.674754405311069 1
0.690551352016233 1
0.706718127392749 1
0.723263389648354 1
0.740195999691564 1
0.757525025877191 1
0.775259748862946 1
0.793409666579749 1
0.811984499318401 1
0.83099419493534 1
0.850448934180268 1
0.870359136148516 1
0.890735463861044 1
0.911588829975082 1
0.932930402628469 1
0.954771611420806 1
0.97712415353465 1
1 1
};
\addplot [semithick, color1, const plot mark right]
table {%
0.01 946
0.0102341140210545 946
0.0104737089795945 946
0.0107189131920513 946
0.0109698579789238 946
0.0112266777351081 946
0.0114895100018731 946
0.0117584955405216 946
0.0120337784077759 946
0.0123155060329283 946
0.0126038292967973 946
0.0128989026125331 946
0.0132008840083142 946
0.0135099352119803 946
0.0138262217376466 946
0.0141499129743458 946
0.0144811822767453 946
0.0148202070579886 946
0.0151671688847092 946
0.0155222535742705 946
0.0158856512942805 946
0.0162575566644379 946
0.0166381688607613 946
0.017027691722259 946
0.0174263338600965 946
0.0178343087693191 946
0.0182518349431904 946
0.0186791359902078 946
0.019116440753857 946
0.0195639834351706 946
0.0200220037181558 946
0.0204907468981585 946
0.0209704640132323 946
0.021461411978584 946
0.0219638537241655 946
0.0224780583354873 946
0.0230043011977292 946
0.0235428641432242 946
0.0240940356023952 946
0.024658110758226 946
0.0252353917043477 946
0.0258261876068268 946
0.0264308148697411 946
0.0270495973046313 946
0.0276828663039207 946
0.0283309610183932 946
0.0289942285388288 946
0.0296730240818887 946
0.0303677111803546 946
0.0310786618778201 946
0.0318062569279412 946
0.0325508859983506 946
0.0333129478793467 946
0.0340928506974681 946
0.0348910121340677 946
0.0357078596490046 946
0.0365438307095726 946
0.037399373024788 946
0.0382749447851631 946
0.0391710149080926 946
0.0400880632889846 946
0.0410265810582719 946
0.0419870708444391 946
0.0429700470432084 946
0.0439760360930272 946
0.045005576757005 946
0.0460592204114511 946
0.0471375313411672 946
0.0482410870416537 946
0.04937047852839 946
0.0505263106533568 946
0.0517092024289676 946
0.0529197873595844 946
0.0541587137807947 946
0.0554266452066311 946
0.0567242606849198 946
0.058052255160949 946
0.0594113398496503 946
0.0608022426164942 946
0.0622257083673023 946
0.0636824994471859 946
0.0651733960488242 946
0.0666991966303012 946
0.0682607183427239 946
0.0698587974678525 946
0.0714942898659758 946
0.073168071434272 946
0.0748810385759002 946
0.0766341086800746 946
0.0784282206133768 946
0.0802643352225717 946
0.0821434358491943 630
0.0840665288561833 540
0.086034644166845 486
0.0880488358164346 432
0.0901101825166502 396
0.0922197882333433 374
0.0943787827777538 342
0.096588322411587 306
0.0988495904662559 288
0.101163797976621 270
0.103532184329566 255
0.105956017927762 238
0.108436596868961 221
0.110975249641207 208
0.113573335834311 187
0.116232246867985 170
0.118953406737032 160
0.121738272773966 140
0.124588336429501 130
0.127505124071301 117
0.13049019780144 104
0.13354515629299 96
0.136671635646201 88
0.139871310264724 80
0.143145893752348 70
0.146497139830729 63
0.149926843278605 56
0.153436840893001 49
0.157029012472938 48
0.160705281826164 42
0.164467617799466 36
0.168318035333096 35
0.172258596539879 28
0.176291411809595 28
0.180418640939207 24
0.184642494289554 20
0.188965233969121 20
0.193389175045523 16
0.197916686785356 16
0.202550193923067 16
0.207292177959537 12
0.212145178491063 12
0.21711179456945 12
0.222194686093952 12
0.227396575235793 8
0.232720247896041 8
0.238168555197616 8
0.243744415012222 6
0.249450813523032 6
0.255290806823952 6
0.261267522556333 6
0.267384161583995 6
0.273643999707467 6
0.280050389418363 6
0.286606761694825 5
0.293316627839004 4
0.300183581357559 3
0.307211299886176 3
0.31440354715915 3
0.321764175025074 3
0.329297125509715 3
0.337006432927193 3
0.344896226040576 2
0.352970730273065 2
0.361234269970943 2
0.369691270719503 2
0.378346261713193 2
0.387203878181256 2
0.396268863870148 2
0.405546073584083 2
0.415040475785048 2
0.42475715525369 2
0.434701315812502 2
0.444878283112759 2
0.455293507486695 2
0.465952566866468 2
0.476861169771447 2
0.488025158365443 2
0.499450511585514 2
0.511143348344017 2
0.523109930805626 2
0.535356667741072 2
0.547890117959394 2
0.560716993820546 2
0.57384416483024 2
0.587278661318948 2
0.601027678207038 2
0.61509857885805 2
0.629498899022189 2
0.644236350872137 2
0.659318827133355 2
0.674754405311069 2
0.690551352016233 2
0.706718127392749 2
0.723263389648354 2
0.740195999691564 2
0.757525025877191 2
0.775259748862946 2
0.793409666579749 2
0.811984499318401 2
0.83099419493534 2
0.850448934180268 2
0.870359136148516 2
0.890735463861044 2
0.911588829975082 2
0.932930402628469 2
0.954771611420806 2
0.97712415353465 2
1 2
};
\end{axis}

\end{tikzpicture}
	\end{subfigure}
	\begin{subfigure}[t]{0.65\columnwidth}
\begin{tikzpicture}

\definecolor{color1}{rgb}{0.12156862745098,0.466666666666667,0.705882352941177}
\definecolor{color0}{rgb}{1,0.498039215686275,0.0549019607843137}

\begin{axis}[
width=7cm,
height=4.5cm,
legend cell align={left},
legend style={draw=white!80.0!black},
tick align=outside,
tick pos=left,
x grid style={white!69.01960784313725!black},
xlabel={\# Components ($n_c$)},
xmin=0, xmax=110,
xtick style={color=black},
y grid style={white!69.01960784313725!black},
ylabel={NRMSE},
title={(c) Validation on Pouring Dataset},
ylabel style={font=\scriptsize, yshift=-10pt},
yticklabel style ={font=\scriptsize},
xlabel style={font=\small, yshift=1.25pt},
xticklabel style ={font=\scriptsize},
scaled y ticks=base 10:2,
xmajorgrids,
ymajorgrids,
ymin=0.00570321434126315, ymax=0.109657816093269,
ytick style={color=black}
]
\addplot [semithick, green!50.0!black]
table {%
0 0.104932606922723
1 0.0910337815117997
2 0.0862611168552077
3 0.0776232407811478
4 0.0721681855099538
5 0.0671040077201177
6 0.0627363002358185
7 0.0595850121409709
8 0.0534118251084207
9 0.050813794008807
10 0.0498196952108898
11 0.0469802000905527
12 0.045289080206288
13 0.0443559149120061
14 0.0442767549317863
15 0.0440169482867722
16 0.0412135868746055
17 0.0408823376660913
18 0.0393495261578208
19 0.0390465310990783
20 0.0380394834109337
21 0.0373250148975021
22 0.0367679803028529
23 0.0359391445495877
24 0.0355791325851553
25 0.0352646032217158
26 0.0349507705138853
27 0.0346456676701622
28 0.0344604160592612
29 0.034169691550158
30 0.0340083953619032
31 0.0339022408792622
32 0.0332849680302991
33 0.0331405175451921
34 0.0327493407610795
35 0.0325524465088783
36 0.0323951348043796
37 0.0321067794134375
38 0.0319621888287532
39 0.0317626594550972
40 0.0315188175815433
41 0.0311808336861624
42 0.0309465504772332
43 0.0307000587576494
44 0.0305822371360089
45 0.0304176303681322
46 0.0301283334970467
47 0.0299624626364963
48 0.0297663556886635
49 0.0294504319071154
50 0.0292687637430081
51 0.029142264790857
52 0.0289449510215104
53 0.0286445808122059
54 0.0283615339921128
55 0.0282720133785334
56 0.0278594310073496
57 0.0277669067415786
58 0.0276254496173647
59 0.0275172649487022
60 0.0272386940553581
61 0.0269951448169198
62 0.0268206803299417
63 0.0266671687345334
64 0.026318250084912
65 0.026103648611458
66 0.0258794995346698
67 0.0255801647767008
68 0.0254367906086762
69 0.025352240622366
70 0.0253106224614436
71 0.0251218649517829
72 0.0250599129904229
73 0.0248895278239849
74 0.0246945055688246
75 0.0246131030628896
76 0.024409014985219
77 0.0242826571678522
78 0.024217637285716
79 0.024139947262001
80 0.0238651845928524
81 0.0237077612694452
82 0.0236254565759266
83 0.0235639571311464
84 0.0233845534299395
85 0.0233161406925455
86 0.0232340342504199
87 0.0231386321351337
88 0.0230069595956385
89 0.0229511618572
90 0.0229076384940499
91 0.0228524138391917
92 0.0227251134568417
93 0.0226363024453711
94 0.0225541650581143
95 0.0225020320521799
96 0.0224187503716626
97 0.0223075998903978
98 0.0222694058479978
99 0.0221980829094481
100 0.0221445169025276
101 0.0221189074575843
102 0.0220957344015882
103 0.0220354649904849
104 0.0220110579581714
105 0.021998215763774
106 0.0219892446127661
107 0.0219723714585089
108 0.0219652746288081
109 0.0219534476657819
};
\addplot [semithick, violet]
table {%
0 0.0937019975341703
1 0.0760561986125083
2 0.0651864203216885
3 0.0551049252696544
4 0.0473467152722209
5 0.0410960588275363
6 0.0355366690750187
7 0.0311742539365687
8 0.0272387556473589
9 0.024368280016745
10 0.022253284937897
11 0.0202499834244811
12 0.0187074188490201
13 0.0172726364974775
14 0.0159738927074504
15 0.014648967457907
16 0.013537104435665
17 0.0126442262205326
18 0.0118602337356423
19 0.0112546476176371
20 0.0107726354433211
21 0.0104284235118089
22 0.0104284235118089
23 0.0104284235118089
24 0.0104284235118089
25 0.0104284235118089
26 0.0104284235118089
27 0.0104284235118089
28 0.0104284235118089
29 0.0104284235118089
30 0.0104284235118089
31 0.0104284235118089
32 0.0104284235118089
33 0.0104284235118089
34 0.0104284235118089
35 0.0104284235118089
36 0.0104284235118089
37 0.0104284235118089
38 0.0104284235118089
39 0.0104284235118089
40 0.0104284235118089
41 0.0104284235118089
42 0.0104284235118089
43 0.0104284235118089
44 0.0104284235118089
45 0.0104284235118089
46 0.0104284235118089
47 0.0104284235118089
48 0.0104284235118089
49 0.0104284235118089
50 0.0104284235118089
51 0.0104284235118089
52 0.0104284235118089
53 0.0104284235118089
54 0.0104284235118089
55 0.0104284235118089
56 0.0104284235118089
57 0.0104284235118089
58 0.0104284235118089
59 0.0104284235118089
60 0.0104284235118089
61 0.0104284235118089
62 0.0104284235118089
63 0.0104284235118089
64 0.0104284235118089
65 0.0104284235118089
66 0.0104284235118089
67 0.0104284235118089
68 0.0104284235118089
69 0.0104284235118089
70 0.0104284235118089
71 0.0104284235118089
72 0.0104284235118089
73 0.0104284235118089
74 0.0104284235118089
75 0.0104284235118089
76 0.0104284235118089
77 0.0104284235118089
78 0.0104284235118089
79 0.0104284235118089
80 0.0104284235118089
81 0.0104284235118089
82 0.0104284235118089
83 0.0104284235118089
84 0.0104284235118089
85 0.0104284235118089
86 0.0104284235118089
87 0.0104284235118089
88 0.0104284235118089
89 0.0104284235118089
90 0.0104284235118089
91 0.0104284235118089
92 0.0104284235118089
93 0.0104284235118089
94 0.0104284235118089
95 0.0104284235118089
96 0.0104284235118089
97 0.0104284235118089
98 0.0104284235118089
99 0.0104284235118089
100 0.0104284235118089
101 0.0104284235118089
102 0.0104284235118089
103 0.0104284235118089
104 0.0104284235118089
105 0.0104284235118089
106 0.0104284235118089
107 0.0104284235118089
108 0.0104284235118089
109 0.0104284235118089
};
\end{axis}

\end{tikzpicture}
	\end{subfigure}
	\vspace{-0.2cm}
	\centering
	\vspace{-0.25cm}
	\begin{tikzpicture}

\definecolor{color0}{rgb}{1,0.647058823529412,0}
\definecolor{color1}{rgb}{0.117647058823529,0.564705882352941,1}

\begin{axis}[
height=2cm,
width=8cm,
hide axis,
xmin=10,
xmax=50,
ymin=0,
ymax=0.5,
legend columns=-1,
legend entries={{CPCA},{PriMos},{PriMos (training set)}, {PriMos (validation set)}},
legend style={at={(0.25,0.1)}, anchor=north, draw=none, font=\small, column sep=2ex, line width=3 pt},
]

\addlegendimage{no markers, color0}
\addlegendimage{no markers, color1}
\addlegendimage{no markers, violet}
\addlegendimage{no markers, green!50.19607843137255!black}

\end{axis}

\end{tikzpicture}
	\caption{The number of parameters (y-axis) needed to achieve a certain accuracy (x-axis) (a), (b). As can be seen PriMos achieve better results with less samples. The error has been normalized in order to highlight the difference in complexity between the two datasets. In plot (b), both PriMo and CPCA fail in reaching a NRMSE lower than $0.08$. This is due to the complexity of the movement in the dataset, and by the maximum number of radial basis functions (20). In plots (c) we observe the comparison between train (blue) and validation error (green) achieved by PriMos in the pouring datasets.}
	\label{fig:analysis}
\end{figure*}

We want to compare the dimensionality reduction in parameter space w.r.t. in configuration space. Therefore, we test the two approaches on two different scenario: the reconstruction of highly uncorrelated movements of a human subject, and on a dataset of pouring movements shown with a 7 DoF robotic arm.
We also perform a qualitative analysis showing how our proposed method can achieve similar results to standard techniques, dramatically reducing the number of parameters.

\paragraph{The Human Motion Dataset}
We want to apply \mbox{PriMos} to reconstruct some human motions contained in the \texttt{MoCap} dataset. The \texttt{MoCap} database contains a wide range of human motions (such as running, walking, picking up boxes, and even dancing) using different subjects (Fig.~\ref{fig:frames}). Human motion is tracked using 41 markers with a Vicon optical tracking system. The data is preprocessed using Vicon Bodybuilder in order to reconstruct a schematic representation of the human body and the relative joint's angles including the system 3D system reference, for a total of 62 values. 
The human motion is known to be highly redundant (i.e. many configurations are highly correlated), and therefore this is in principle the best case for the dimensionality-reduction in the configuration-space. The presence of many different typologies of movements, makes the application of PriMos challenging since our algorithm relies on the correlation between movements.
In our experiments, we use the 42 movements recorded for the subject \#143.
\begin{figure*}[t]
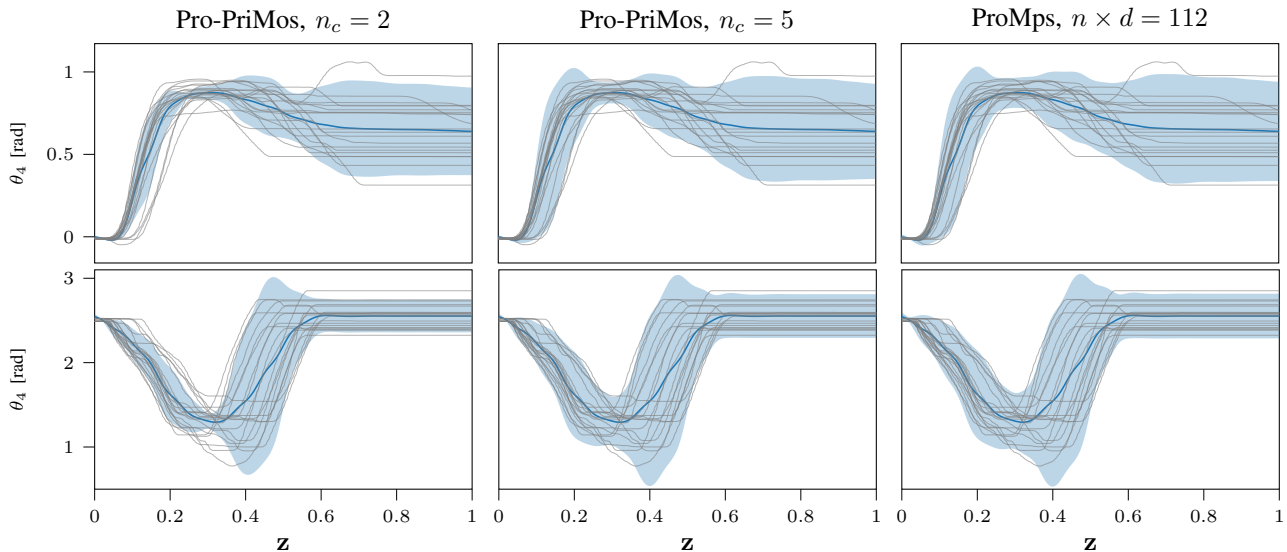

	\begin{subfigure}[t]{0.75\columnwidth}
		\input{plots/variance/variance_primo_2_0.tikz}
	\end{subfigure}
	\begin{subfigure}[t]{0.6\columnwidth}
		\input{plots/variance/variance_primo_5_0.tikz}
	\end{subfigure}
	\hspace{-2pt}
	\begin{subfigure}[t]{0.6\columnwidth}
		\input{plots/variance/variance_promp_5_0.tikz}
	\end{subfigure}
	
	\begin{subfigure}[t]{0.73\columnwidth}
		\input{plots/variance/variance_primo_2_4.tikz}
	\end{subfigure}
	\begin{subfigure}[t]{0.6\columnwidth}
		\input{plots/variance/variance_primo_5_4.tikz}
	\end{subfigure}
	\hspace{-2pt}
	\begin{subfigure}[t]{0.6\columnwidth}
		\input{plots/variance/variance_promp_5_4.tikz}
	\end{subfigure}
\vspace{-0.5em}
	\caption{Mean movement of the joints most involved in the movements (shoulder and elbow), accompanied by the standard deviation ($\pm 2\sigma$) for Pro-PriMos with 2 and 5 components, and with ProMPs with 112 parameters for the mean. Note that the covariance matrix is encoded with 4, 25 and 12544 values respectively.}
	\label{fig:variance}
\end{figure*}
\paragraph{The Pouring Task}
In the pouring task, we use a KUKA light-weight robotic arm with 7 DoF accompanied with a DLR-hand as an end-effector to pour some ``liquid'' (which, for safety reasons, is replaced by granular sugar). We record some motions from a human demonstrator, pouring some sugar in a bowl. The motion is recorded setting the robotic arm in kinestatic teaching. The quantity of sugar contained in the bowl is recorded by a DYMO digital scale with a sensitivity of $2g$. We aim to reconstruct the movements and understand whether our method is able to pour a similar amount of sugar: this experiment gives us a qualitative understanding, beyond the numerical accuracy analysis, to investigate the effectiveness of our algorithm in a real robot-learning task. 

\subsection{Accuracy of Movement Reconstruction}
We want to compare the quality of the reconstruction using PriMos, and the dimensionality reduction in the configuration space. In order to be fair, we use PCA as reduction technique for both methods. In the remainder of the paper, we refer to the dimensionality reduction in the configuration space with the acronym ``CPCA''. 

\paragraph{Parameters vs Error Analysis}
In this experiment, we want to assess the number of parameters needed to obtain a certain accuracy in the reconstruction. We, therefore, apply PriMos  and CPCA to both the \texttt{MoCap} dataset and a dataset containing 15 different pouring movements. The CPCA admits several configurations for a fixed number of parameters (e.g., $100$ parameters can be obtained using $2$ components of the PCA and $50$ features or $4$ components and $25$ features, and so on). Therefore we created a grid of configuration in order to obtain the minimal amount of parameters to assess a certain accuracy. 
In order to make the results comparable between the \texttt{MoCap} and the pouring datasets, we use the \textsl{normalized root mean square error} (NRMSE) where the normalization is achieved dividing the RMSE by the standard deviation of the dataset.
Fig.~\ref{fig:analysis} depicts the NRMSE of PriMos and CPCA for the \texttt{MoCap} and pouring datasets. In both cases, the reduction in parameter space performed by PriMos requires significantly fewer parameter than the one performed in joint space by CPCA. Notably, for the \texttt{MoCap} database, the maximum number of basis functions chosen is not sufficient to achieve less than the $0.08$ reconstruction error.

\paragraph{Leave-One-Out Analysis}
One might argue that our method ``memorizes'' the movements contained in the datasets. We want therefore to inspect the ability of PriMos to reconstruct movements not contained in the dataset (in other words, we want to understand if our method \textsl{generalizes}). We achieve this analysis using an increasing number of parameters and plotting the NRMSE using a leave-one-out strategy in the pouring dataset (we train on all the possible $n$ subsets of $n-1$ movements, and we test on the movement left out). More precisely, we fit $\overline{\bm{\omega}}$ and $\bm{\Omega}$ using the training set and then we test the error in the validation set. Fig.~\ref{fig:analysis} shows that the error in the validation sets is not significantly higher than the training-error. Interestingly, while the training error becomes almost constant after $20$ components, the validation error keeps getting lower. This behavior suggests that our approach is robust against overfitting.

\subsection{A Qualitative Evaluation}
We want to understand the applicability of PriMos to real robotics. The study of the accuracy conducted is important, but on its own, it does not give us a feeling about how the method works in practice. For this reason, we use the 15 trajectories contained in the pouring dataset, and we measured the quantity of sugar poured in the bowl.

\paragraph{Single Movement Reconstruction}

\begin{figure}[h]
	\begin{subfigure}[t]{0.48\columnwidth}
\begin{tikzpicture}
\definecolor{color0}{rgb}{0.12156862745098,0.466666666666667,0.705882352941177}
\definecolor{color1}{rgb}{1,0.498039215686275,0.0549019607843137}
\begin{axis}[
height=6cm,
width=8.5cm,
legend cell align={left},
legend style={at={(0.5,0.25)}, anchor=north west, draw=white!80.0!black},
tick align=outside,
tick pos=left,
x grid style={lightgray!92.0261437908!black},
xmin=-7.4116227326785, xmax=155.644077386248,
xtick style={color=black},
xlabel={Demonstrated Weight ($g$)},
y grid style={lightgray!92.0261437908!black},
ymin=-7.51557318506185, ymax=157.827036886299,
ytick style={color=black},
ylabel={Reconstructed Weight ($g$)},
ylabel style={yshift=-5pt},
yticklabel style = {font=\scriptsize},
xticklabel style = {font=\scriptsize}
]
\addplot [only marks, draw=blue, fill=color0, colormap/viridis, opacity=0.75]
table{%
x                      y
66 62
12 12
96 126
40 22
112 108
126 124
28 10
66 112
148 150
36 46
84 88
134 118
74 80
50 60
104 98
66 62
12 14
96 118
40 22
112 102
126 118
28 8
66 108
148 148
36 46
84 82
134 110
74 76
50 58
104 92
66 66
12 16
96 128
40 20
112 108
126 124
28 10
66 112
148 148
36 44
84 88
134 116
74 76
50 56
104 100
};
\addlegendentry{PriMos (5 params)}
\addplot [only marks, draw=red, fill=color1, colormap/viridis, opacity=0.75]
table{%
x                      y
66 58
12 6
96 96
40 42
112 120
126 144
28 24
66 66
148 148
36 34
84 82
134 144
74 74
50 54
104 106
66 62
12 10
96 92
40 42
112 114
126 134
28 22
66 64
148 148
36 36
84 84
134 138
74 72
50 50
104 104
66 64
12 14
96 98
40 40
112 122
126 144
28 22
66 66
148 148
36 34
84 86
134 144
74 76
50 52
104 110
};
\addlegendentry{MPs (112 params)}
\addplot [semithick, green!50.0!black, forget plot]
table {%
0 0
148 148
};
\end{axis}

\end{tikzpicture}
	\end{subfigure}
\vspace{-1em}
	\caption{Precision of the reconstructed weight. The diagonal line represents the best possible reconstruction. For each demonstration, we repeated three times the reconstructed movement in order. \label{fig:reconstruction}}
\end{figure}
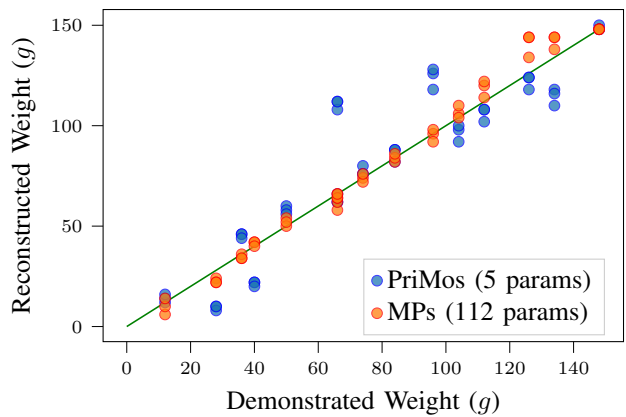

We run all the 15 trajectories reconstructed with the classic MPs and with PriMos for three times in order to average the stochasticity inherent to the experiment (a slight perturbation of the glass position or of the sugar contained in the glass might perturb the resulting quantity of sugar poured). A video of the demonstration as well of the reconstruction is available as supplementary material. Fig.~\ref{fig:reconstruction} represents a confusion scatter plot: on the $x$-axis we have the weight of the sugar observed during the demonstration, while on the $y$-axis we observe the reconstructed movement both with MPs (112 parameters) and with PriMos (5 parameters). The ideal situation is when the points lye down on the identity line shown in green (perfect reconstruction). We observe that PriMos, except few outliers, reaches a similar accuracy to the MPs, but with $4.5\%$ of the parameters. 

\paragraph{Probabilistic Representation}
An important aspect of our framework is the possibility to represent the \textsl{distribution} of movements. For this reason, we show novel movements generated both with ProMPs and Pro-PriMos given $25$ demonstrations. 
The approximation of the full covariance-matrix is usually demanding, as it scales quadratically w.r.t. the number of parameters encoding the mean movement. Fig.~\ref{fig:variance} depicts the standard deviation of the shoulder of the robotic arm, within the demonstrated movement; Pro-PriMos seems to represent the stochasticity very similarly to the ProMPs, although the covariance matrix in the first case requires $5\times 5$ values, in contrast to $112 \times 112$  required by the ProMPs. We also note that the variability of the Pro-PriMos seems to be lower than the ProMPs' one: this behavior is explained by the fact that for a fixed set of features, PriMos represents a subset of movements of ProMPs.
We furthermore sample 40 movements both from ProMPs and Pro-PriMos with $n_c=5$, and we measure the quantity of sugar poured. Table~\ref{tab:pouring} shows that both the methods are able to pour all ranges of sugar (even if with different proportion from the demonstrated data). However, ProMPs occasionally fails to generate a good movement, and it pours the sugar outside the bowl. The video in the Supplementary Materials shows this situation. 
\begin{table}[t]
	\centering
	\begin{tabular}{| l | r r r r r|}\rowcolor{gray!50}
		\hline
		 Method & 0-40$g$ & 40-80$g$ & 80-120$g$ & 120-150$g$ & Failure \\
		 \hline
		\rowcolor{white}
		Demonstrations & 20\% & 33\% & 27\% & 20\% & 0\% \\
		\rowcolor{gray!25}
		ProMPs & 43\% & 7\% & 20\% & 17\% & 13\%\\
		\rowcolor{white}
		Pro-PriMos & 53\% & 20\% & 3\% & 24\% & 0\\
		\hline
	\end{tabular}
\caption{Measurement of sugar poured from 40 movements sampled both from ProMPs and Pro-PriMos. The methods manage to roughly resemble the original distribution, however ProMPs exhibit unwanted behaviors, like pouring most of the sugar outside the bowl (classified as ``failures'').\label{tab:pouring}}
\end{table}

\section{Conclusion}
The main contribution of this paper is the analysis of the dimensionality reduction in parameter space in the context of MPs.
Our findings suggest that this reduction is more efficient than the one in the configuration space. The novel approach (PriMos), which operates dimensionality reduction in the parameter space using the principal component analysis, is enriched with a probabilistic treatment of the parameters in order to inherit all the convenient properties of the ProMPs.
We tested our approach both on a robotic task as well as in a challenging dataset of human movements.
Our method compares well against the dimensionality reduction in configuration space, and exhibits a significant reduction of parameters with a modest loss of accuracy, even in the probabilistic setting.
We argue that these insights are helpful to develop more efficient robot learning techniques.

As future work, we will investigate the application of different dimensionality reduction techniques in parameter space, with a special focus in the RL context.

%

\section*{ACKNOWLEDGMENT}
The research is financially supported by the Bosch-Forschungsstiftung program and by SKILLS4ROBOT under grant agreement \#640554. We would like to thank Georgia Chalvatzaki for her useful tips.
\bibliographystyle{IEEEtran}
\bibliography{zotero}

\begin{thebibliography}{10}
\providecommand{\url}[1]{#1}
\csname url@rmstyle\endcsname
\providecommand{\newblock}{\relax}
\providecommand{\bibinfo}[2]{#2}
\providecommand\BIBentrySTDinterwordspacing{\spaceskip=0pt\relax}
\providecommand\BIBentryALTinterwordstretchfactor{4}
\providecommand\BIBentryALTinterwordspacing{\spaceskip=\fontdimen2\font plus
\BIBentryALTinterwordstretchfactor\fontdimen3\font minus
  \fontdimen4\font\relax}
\providecommand\BIBforeignlanguage[2]{{%
\expandafter\ifx\csname l@#1\endcsname\relax
\typeout{** WARNING: IEEEtran.bst: No hyphenation pattern has been}%
\typeout{** loaded for the language `#1'. Using the pattern for}%
\typeout{** the default language instead.}%
\else
\language=\csname l@#1\endcsname
\fi
#2}}

\bibitem{mnih_human-level_2015}
\BIBentryALTinterwordspacing
V.~Mnih, K.~Kavukcuoglu, D.~Silver, A.~A. Rusu, J.~Veness, M.~G. Bellemare,
  A.~Graves, M.~Riedmiller, A.~K. Fidjeland, G.~Ostrovski, S.~Petersen,
  C.~Beattie, A.~Sadik, I.~Antonoglou, H.~King, D.~Kumaran, D.~Wierstra,
  S.~Legg, and D.~Hassabis, ``\BIBforeignlanguage{en}{Human-{Level} {Control}
  {Through} {Deep} {Reinforcement} {Learning}},''
  \emph{\BIBforeignlanguage{en}{Nature}}, vol. 518, no. 7540, pp. 529--533,
  2015. [Online]. Available: \url{http://www.nature.com/articles/nature14236}
\BIBentrySTDinterwordspacing

\bibitem{lillicrap_continuous_2016}
\BIBentryALTinterwordspacing
T.~P. Lillicrap, J.~J. Hunt, A.~Pritzel, N.~Heess, T.~Erez, Y.~Tassa,
  D.~Silver, and D.~Wierstra, ``Continuous {Control} with {Deep}
  {Reinforcement} {Learning},'' in \emph{International {Conference} on
  {Learning} {Representations}}, 2016, arXiv: 1509.02971. [Online]. Available:
  \url{http://arxiv.org/abs/1509.02971}
\BIBentrySTDinterwordspacing

\bibitem{schulman_proximal_2017}
J.~Schulman, F.~Wolski, P.~Dhariwal, A.~Radford, and O.~Klimov, ``Proximal
  {Policy} {Optimization} {Algorithms},'' \emph{arXiv preprint
  arXiv:1707.06347}, 2017.

\bibitem{schaal_is_1999}
S.~Schaal, ``Is {Imitation} {Learning} the {Route} to {Humanoid} {Robots}?''
  \emph{Trends in cognitive sciences}, vol.~3, no.~6, pp. 233--242, 1999.

\bibitem{billard_robot_2004}
A.~Billard and R.~Siegwart, ``Robot {Learning} from {Demonstration},''
  \emph{Robotics and Autonomous Systems}, vol.~2, no.~47, pp. 65--67, 2004.

\bibitem{argall_survey_2009}
B.~D. Argall, S.~Chernova, M.~Veloso, and B.~Browning, ``A {Survey} of {Robot}
  {Learning} from {Demonstration},'' \emph{Robotics and autonomous systems},
  vol.~57, no.~5, pp. 469--483, 2009.

\bibitem{rana_towards_2018}
M.~Rana, M.~Mukadam, S.~R. Ahmadzadeh, S.~Chernova, and B.~Boots, ``Towards
  {Robust} {Skill} {Generalization}: {Unifying} {Learning} from {Demonstration}
  and {Motion} {Planning},'' in \emph{Intelligent robots and systems}, 2018.

\bibitem{davella_combinations_2003}
A.~D'Avella, P.~Saltiel, and E.~Bizzi, ``Combinations of {Muscle} {Synergies}
  in the {Construction} of a {Natural} {Motor} {Behavior},'' \emph{Nature
  Neuroscience}, vol.~6, no.~3, pp. 300--308, 2003.

\bibitem{schaal_learning_2005}
S.~Schaal, J.~Peters, J.~Nakanishi, and A.~Ijspeert, ``Learning {Movement}
  {Primitives},'' in \emph{Robotics research. {The} eleventh international
  symposium}.\hskip 1em plus 0.5em minus 0.4em\relax Springer, 2005, pp.
  561--572.

\bibitem{khansari-zadeh_learning_2011}
S.~M. Khansari-Zadeh and A.~Billard, ``Learning {Stable} {Nonlinear}
  {Dynamical} {Systems} with {Gaussian} {Mixture} {Models},'' \emph{IEEE
  Transactions on Robotics}, vol.~27, no.~5, pp. 943--957, 2011.

\bibitem{paraschos_probabilistic_2013}
A.~Paraschos, C.~Daniel, J.~Peters, and G.~Neumann, ``Probabilistic {Movement}
  {Primitives},'' in \emph{Advances in {Neural} {Information} {Processing}
  {Systems} ({NIPS})}.\hskip 1em plus 0.5em minus 0.4em\relax mit press, 2013.

\bibitem{amor_interaction_2014}
H.~B. Amor, G.~Neumann, S.~Kamthe, O.~Kroemer, and J.~Peters, ``Interaction
  {Primitives} for {Human}-{Robot} {Cooperation} {Tasks},'' in \emph{{IEEE}
  international conference on robotics and automation ({ICRA})}.\hskip 1em plus
  0.5em minus 0.4em\relax IEEE, 2014, pp. 2831--2837.

\bibitem{maeda_learning_2014}
G.~Maeda, M.~Ewerton, R.~Lioutikov, H.~B. Amor, J.~Peters, and G.~Neumann,
  ``Learning {Interaction} for {Collaborative} {Tasks} with {Probabilistic}
  {Movement} {Primitives},'' in \emph{{IEEE}-{RAS} {International} {Conference}
  on {Humanoid} {Robots}}.\hskip 1em plus 0.5em minus 0.4em\relax IEEE, 2014,
  pp. 527--534.

\bibitem{koert_demonstration_2016}
D.~Koert, G.~Maeda, R.~Lioutikov, G.~Neumann, and J.~Peters, ``Demonstration
  {Based} {Trajectory} {Optimization} for {Generalizable} {Robot} {Motions},''
  in \emph{{IEEE}-{RAS} 16th {International} {Conference} on {Humanoid}
  {Robots} ({Humanoids})}.\hskip 1em plus 0.5em minus 0.4em\relax IEEE, 2016,
  pp. 515--522.

\bibitem{maeda_active_2017}
G.~Maeda, M.~Ewerton, T.~Osa, B.~Busch, and J.~Peters, ``Active {Incremental}
  {Learning} of {Robot} {Movement} {Primitives},'' 2017.

\bibitem{stark_experience_2019}
S.~Stark, J.~Peters, and E.~Rueckert, ``Experience {Reuse} with {Probabilistic}
  {Movement} {Primitives},'' \emph{arXiv preprint arXiv:1908.03936}, 2019.

\bibitem{peters_reinforcement_2008}
J.~Peters and S.~Schaal, ``Reinforcement {Learning} of {Motor} {Skills} with
  {Policy} {Gradients},'' \emph{Neural networks}, vol.~21, no.~4, pp. 682--697,
  2008.

\bibitem{kober_policy_2009}
J.~Kober and J.~R. Peters, ``Policy {Search} for {Motor} {Primitives} in
  {Robotics},'' in \emph{Advances in {Neural} {Information} {Processing}
  {Systems}}, 2009, pp. 849--856.

\bibitem{mulling_learning_2013}
K.~M{\"u}lling, J.~Kober, O.~Kroemer, and J.~Peters, ``Learning to {Select} and
  {Generalize} {Striking} {Movements} in {Robot} {Table} {Tennis},'' \emph{The
  International Journal of Robotics Research}, vol.~32, no.~3, pp. 263--279,
  2013.

\bibitem{paraschos_using_2018}
A.~Paraschos, C.~Daniel, J.~Peters, and G.~Neumann, ``Using {Probabilistic}
  {Movement} {Primitives} in {Robotics},'' \emph{Autonomous Robots (AURO)},
  no.~3, pp. 529--551, 2018.

\bibitem{colome_dimensionality_2014-1}
A.~Colom{\'e} and C.~Torras, ``Dimensionality {Reduction} and {Motion}
  {Coordination} in {Learning} {Trajectories} with {Dynamic} {Movement}
  {Primitives},'' in \emph{{IEEE}/{RSJ} {International} {Conference} on
  {Intelligent} {Robots} and {Systems}}.\hskip 1em plus 0.5em minus 0.4em\relax
  IEEE, 2014, pp. 1414--1420.

\bibitem{colome_dimensionality_2014}
A.~Colom{\'e}, G.~Neumann, J.~Peters, and C.~Torras, ``Dimensionality
  {Reduction} for {Probabilistic} {Movement} {Primitives},'' in
  \emph{International {Conference} on {Humanoid} {Robots}}.\hskip 1em plus
  0.5em minus 0.4em\relax IEEE, 2014, pp. 794--800.

\bibitem{chen_efficient_2015}
N.~Chen, J.~Bayer, S.~Urban, and P.~Van Der~Smagt, ``Efficient {Movement}
  {Representation} by {Embedding} {Dynamic} {Movement} {Primitives} in {Deep}
  {Autoencoders},'' in \emph{{IEEE}-{RAS} 15th {International} {Conference} on
  {Humanoid} {Robots} ({Humanoids})}.\hskip 1em plus 0.5em minus 0.4em\relax
  IEEE, 2015, pp. 434--440.

\bibitem{chen_dynamic_2016}
N.~Chen, M.~Karl, and P.~Van Der~Smagt, ``Dynamic {Movement} {Primitives} in
  {Latent} {Space} of {Time}-{Dependent} {Variational} {Autoencoders},'' in
  \emph{2016 {IEEE}-{RAS} 16th {International} {Conference} on {Humanoid}
  {Robots} ({Humanoids})}.\hskip 1em plus 0.5em minus 0.4em\relax IEEE, 2016,
  pp. 629--636.

\bibitem{colome_dimensionality_2018}
A.~Colom{\'e} and C.~Torras, ``Dimensionality {Reduction} for {Dynamic}
  {Movement} {Primitives} and {Application} to {Bimanual} {Manipulation} of
  {Clothes},'' \emph{IEEE Transactions on Robotics}, vol.~34, no.~3, pp.
  602--615, 2018.

\bibitem{rueckert_extracting_2015}
E.~Rueckert, J.~Mundo, A.~Paraschos, J.~Peters, and G.~Neumann, ``Extracting
  {Low}-{Dimensional} {Control} {Variables} for {Movement} {Primitives},'' in
  \emph{2015 {IEEE} {International} {Conference} on {Robotics} and {Automation}
  ({ICRA})}.\hskip 1em plus 0.5em minus 0.4em\relax IEEE, 2015, pp. 1511--1518.

\bibitem{pearson_lines_1901}
K.~Pearson, ``On {Lines} and {Planes} of {Closest} fit to {Systems} of {Points}
  in {Space},'' \emph{The London, Edinburgh, and Dublin Philosophical Magazine
  and Journal of Science}, vol.~2, no.~11, pp. 559--572, 1901.

\bibitem{golub_matrix_2012}
G.~H. Golub and C.~F. Van~Loan, \emph{Matrix {Computations}}, 3rd~ed.\hskip 1em
  plus 0.5em minus 0.4em\relax JHU press, 2012.

\end{thebibliography}

\end{document}